\documentclass[runningheads]{llncs}
\usepackage{amsmath,amssymb} 
\usepackage{makeidx}
\usepackage{graphicx}
\usepackage{color}

\usepackage[ruled,vlined,linesnumbered]{algorithm2e}
\usepackage{algpseudocode}
\usepackage{colortbl}
\usepackage{booktabs} 
\usepackage[colorlinks]{hyperref}

\usepackage{array}
\newcolumntype{L}[1]{>{\raggedright\arraybackslash}p{#1}}
\newcolumntype{C}[1]{>{\centering\arraybackslash}p{#1}}
\newcolumntype{R}[1]{>{\raggedleft\arraybackslash}p{#1}}
\newcommand{\tabsection}[2]{%
    \cmidrule{0-3}
    \multicolumn{4}{c}{\cellcolor[gray]{0.9}

        \textbf{#1} #2
    } \\
    \cmidrule{0-3}
}

\newcommand{\coltabsection}[2]{}

\begin{document}
	\pagestyle{headings}
	\mainmatter

	\def\GCPR19SubNumber{119}

	\title{Semi-Supervised Segmentation of Salt Bodies in Seismic Images using an Ensemble of Convolutional Neural Networks}

	\titlerunning{Semi-Supervised Segmentation of Salt Bodies in Seismic Images using CNNs}
	\authorrunning{Yauhen Babakhin, Artsiom Sanakoyeu, Hirotoshi Kitamura}

    \author{
Yauhen Babakhin\inst{1} \and
Artsiom Sanakoyeu\inst{2} \and
Hirotoshi Kitamura\inst{3}
}

    \institute{
$^1$ H2O.ai, Belarus \\ {\tt\small yauhen.babakhin@h2o.ai} \\
$^2$ Heidelberg Collaboratory for Image Processing, IWR,  Heidelberg University, Germany \\ {\tt\small artsiom.sanakoyeu@iwr.uni-heidelberg.de} \\
$^3$ Ritsumeikan University, Japan \\ {\tt\small ritskitamura@gmail.com}
}

    \maketitle

\begin{abstract}
Seismic image analysis plays a crucial role in a wide range of industrial applications and has been receiving significant attention. One of the essential challenges of seismic imaging is detecting subsurface salt structure which is indispensable for the identification of hydrocarbon reservoirs and drill path planning.
Unfortunately, the exact identification of large salt deposits is notoriously difficult
and professional seismic imaging often requires expert human interpretation of salt bodies. 
Convolutional neural networks (CNNs) have been successfully applied
in many fields, and several attempts have been made in the field of seismic imaging. 
But the high cost of manual annotations by geophysics experts and scarce publicly available labeled datasets hinder the performance of the existing CNN-based methods.
In this work, we propose a semi-supervised method for segmentation (delineation) of salt bodies in seismic images which utilizes unlabeled data for multi-round self-training. To reduce error amplification during self-training we propose a scheme which uses an ensemble of CNNs. We show that our approach outperforms state-of-the-art on the TGS Salt Identification Challenge dataset and is ranked the first among the $3234$ competing methods.
The source code is available at \href{https://github.com/ybabakhin/kaggle_salt_bes_phalanx}{GitHub}.
	\end{abstract}

\section{Introduction}

\begin{figure}[t!]
\begin{center}
\includegraphics[width=0.99\linewidth]{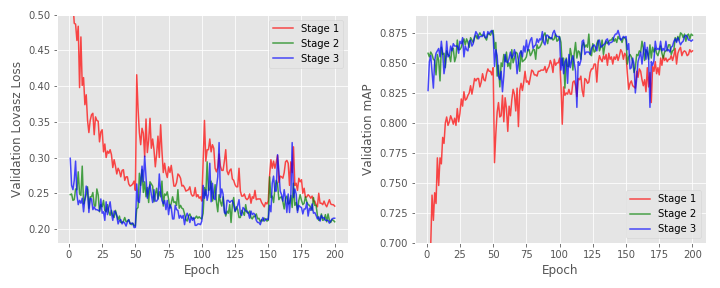}
\end{center}
   \caption{Progress of the validation loss (top) and the validation mAP score (bottom) during training our U-ResNet34 model on TGS Salt Identification Challenge dataset \cite{kaggle_salt_competition} for $K=3$ rounds. Every next round the model converges faster and achieves better local minima. Loss spikes every $50$ epochs correspond to the cycles of the cosine annealing learning rate schedule.}
\label{fig:train_progress_plot}
\end{figure}

\begin{figure}[t!]
\begin{center}
\includegraphics[width=0.68\linewidth]{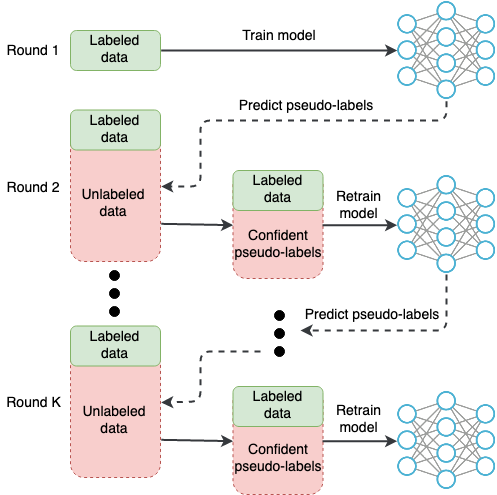}
\end{center}
   \caption{The pipeline of the proposed self-training procedure. We do $K$ rounds of retraining the model. Every round we train the model on the available labeled data and predicted confident pseudo-labels for the unlabeled data. \emph{All} pseudo-labels are recalculated at the end of every round.}
\label{fig:self_training_pipeline}
\end{figure}

One of the major challenges of seismic imaging is localization and delineation of subsurface salt bodies. The precise location of salt deposits helps to identify reservoirs of hydrocarbons, such as crude oil or natural gas, which are trapped by overlying rock-salt formations due to the exceedingly small permeability of the latter. 

\begin{figure*}[t]
\begin{center}
 \includegraphics[width=1\linewidth]{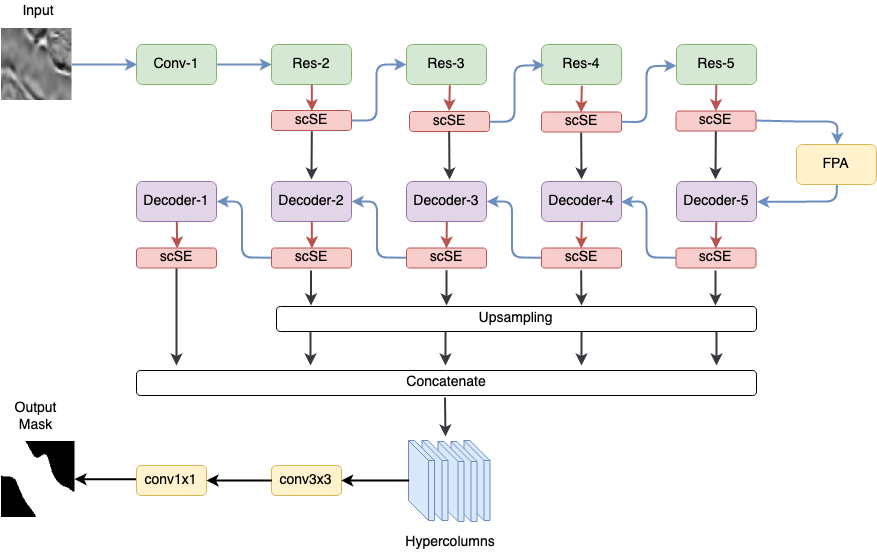} 
\end{center}
   \caption{The outline of the U-ResNet34/U-ResNeXt50 architecture proposed. The difference between U-ResNet34 and U-ResNeXt50 is only in the structure of the encoder blocks (green).
   We insert scSE modules \cite{scse} after each encoder (green) and decoder (purple) blocks. 
  Encoder blocks are connected with the corresponding decoder blocks using skip-connections.
  We use a Feature Pyramid Attention module (FPA) \cite{fpa} after the last encoder block. All outputs of the decoder blocks are upsampled to have the same size as the output of the last decoder bock. Obtained feature maps are concatenated together into hypercolumns \cite{hypercolumns}, which are used for prediction of the segmentation mask after applying two  convolutional layers.}
\label{fig:architecture}
\end{figure*}

\begin{figure*}[t]
\begin{center}
\includegraphics[width=0.9\linewidth]{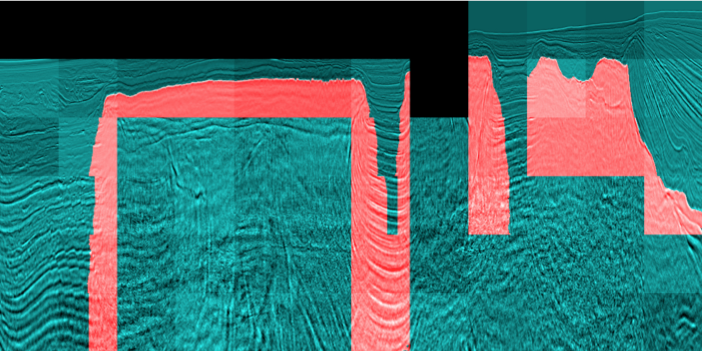} 

\end{center}
   \caption{Example of the 6x12 mosaic of train patches from the TGS Salt Identification Challenge dataset \cite{kaggle_salt_competition}. Each patch is $101 \times 101$ pixels. 
   Green patches denote patches without the salt boundary; green/red patches indicate patches containing the salt body boundary; black color means missing mosaic patches. }
\label{fig:data_example_mosaic}
\end{figure*}

Modern seismic imaging techniques result in large amounts of unlabeled data which have to be interpreted.
Unfortunately, the exact identification of large salt deposits is notoriously difficult \cite{jones2014seismic} and often requires manual interpretation of seismic images by the domain experts. Despite being highly time-consuming and expensive, manual interpretation induces a subjective human bias, which can lead to potentially dangerous situations for oil and gas company drillers. 

In recent years, a number of tools for automatic or semi-automatic seismic interpretation have been proposed \cite{Pitas1992,halpert2008salt,hegazy2014texture,seismic_zhao2015_review,wu2016methods,bedi2018sfa,di2018multi_kmeans,seismic_wrona2018seismic} to speed-up the interpretation process and, to some extent, reduce the human bias. However, these methods do not generalize well for complex cases since they rely on handcrafted features. 

The advent of convolutional neural networks (CNNs) brought significant advancements in different problems and several attempts have been made to apply CNNs in the field of seismic imaging \cite{seismic_cnn_waldeland2018,seismic_Dramsch2018,wang2018automatic_seismic,seismic_salt_cnn_zeng2018automatic}. CNNs overcome the need for manual feature design and show superior performance on the tasks of the salt body delineation compared to the methods based on the handcrafted features. 
However, a low amount of publicly available annotated seismic images hinder the performance of the existing CNN-based methods since CNNs are notoriously hungry for data.

To overcome the shortage of labeled data, we propose a semi-supervised method for segmentation of salt bodies in seismic images which can make use of abundant unlabeled data. The unlabeled images are utilized for self-training \cite{self_labeled_survey}. The proposed self-training procedure (see Fig.~\ref{fig:self_training_pipeline}) is an iterative process which extends the labeled dataset by alternating between training the model and pseudo-labeling (i.e. imputing the labels on the unlabeled data). We do $K$ rounds of retraining the model (see the straining in Fig.~\ref{fig:train_progress_plot}). At the first round, we train model solely on the available labeled data and then predict labels on the unlabeled data. Every next round we use for training both original labeled data and the pseudo-labels obtained at the previous round. 
The error amplification is a well-known problem in self-training \cite{self_training_error_amp} when the error is accumulated during self-training rounds and the models tend to generate less reliable predictions during the time.
To mitigate it we propose to train an ensemble of CNNs and predict labels on the unlabeled data using the average voting of the models in the ensemble. Average voting scheme corrects examples which could be mislabeled by one of the models, hence facilitates more reliable pseudo-labeling. Moreover, to further reduce the error amplification we retrain our models from scratch and predict labels for \emph{all} unlabeled examples every round in similar spirit as \cite{self_training_error_amp}.

We conduct experiments on the largest available to our knowledge dataset for salt body delineation -- TGS Salt Identification Challenge dataset \cite{kaggle_salt_competition}. This dataset was collected by TGS, the world's leading geoscience data company, and was provided in the Kaggle competition. Our approach achieves state-of-the-art performance on this dataset featuring the \emph{first place} in the global ranking among $3234$ competitors.

In summary, the contribution of this work is as follows: (i) we propose an iterative self-training approach for semantic segmentation which benefits from unlabeled data; 
(ii) we build a sophisticated network architecture which is tailored for the task of salt body delineation (see Fig.~\ref{fig:architecture}); (iii) we evaluate our approach on a real-world salt body delineation dataset -- TGS Salt Identification Challenge \cite{kaggle_salt_competition}, where the proposed method achieves the state-of-the-art performance outperforming \emph{all} other competing teams.

\section{Related work}
    A lot of research efforts have been devoted to interpretation of seismic images
    \cite{telford1990applied,Pitas1992,seismic_zhao2015_review,wu2016methods,bedi2018sfa}.
    With the advent of CNNs, several approaches have been proposed for supervised seismic image interpretation using deep learning  \cite{seismic_multiclass_di2018real,seismic_cnn_waldeland2018,seismic_salt_cnn_zeng2018automatic}. But the small size of the available datasets and lack of the annotations seismic image interpretation did not allow to unfold the full potential of the CNNs.
    
    The recent trend in the Computer Vision community is unsupervised or self-supervised learning
    \cite{doersch2015unsupervised,noroozi2016unsupervised,larsson2016learning,cliquecnn,posets,sanakoyeu2018pr,jiang2018self_depth,lee2017unsupervised,buchler2018improving}
    which can make use of abundant unlabeled visual data available on the internet and avoid costly manual annotations. Another class of methods which lies between completely unsupervised methods and supervised methods is semi-supervised learning. It jointly utilizes a large amount of unlabeled data, together with the labeled data \cite{ssl_survey}. The semi-supervised technique most relevant to our work is self-training \cite{yarowsky1995unsupervised,li2005setred,self_labeled_survey}.
    In the self-training, a classifier is trained with an initially small number of labeled examples, then it predicts labels for unlabeled points. After that, the classifier is retrained with its own most confident predictions, together with initially provided labeled examples. However existing self-training approaches \cite{self_train_trees_fazakis2016self,text_progressive_semisup,self_training_xray,informatica_self_labeled_ensemble} are based on hand-crafted features which are much more limited than the features learned by CNNs. 
    \cite{pseudo_label2013} and \cite{wang2017deep_growing} use CNNs in the self-training framework, but they apply it to relatively simple classification datasets like MNIST \cite{mnist} and CIFAR-10 \cite{cifar}. 
    The most relevant self-training approach which is based on CNN features is \cite{han2018semi_remote_sensing}, which is designed for image classification task and uses pretrained CNNs as the fixed feature-extractors while training SVM classifier on top. In contrast, our approach is the first to our knowledge which proposes a self-training procedure for semantic segmentation task and it learns CNN features end-to-end. Moreover, our method reduces the error amplification \cite{self_training_error_amp} by using an ensemble of the networks and by retraining from scratch and recalculating pseudo-labels every training round.
    
    Another work related to ours is \cite{peters2018_horizons}. Authors try to mitigate the high cost of manual annotations of seismic images by introducing an approach which can utilize sparse annotations instead of the commonly used dense segmentation masks.

\section{Method}
The salt body delineation problem can be reduced to the task of semantic image segmentation \cite{seismic_Dramsch2018}, therefore we design our model to predict a binary segmentation mask \cite{chen2014semantic} for the salt body. We will further use the terms segmentation and delineation interchangeably in the text.

In this section, we first present the proposed iterative self-training procedure (Sect.~\ref{sec:self_training}) which can make use of unlabeled samples for training. Then we describe the ensemble used for training and the network architectures in detail (Sect.~\ref{sec:architecture}). 

\subsection{Self-training process}\label{sec:self_training}

Since the labeled data available for the salt body delineation task is scarce, we propose to produce pseudo-labels for unlabeled data and use the pseudo-labels along with ground truth labels to train the model. We refer to this process as self-training. Our self-training procedure is a $K$-round iterative process where each round has 2 steps: (a) training the model using the labeled dataset extended with pseudo-labels; (b) updating pseudo-labels for unlabeled data.

During the first round, we train the model using the ground truth labels only. Then we predict pseudo-labels for all unlabeled data by assigning to each pixel in the image the most probable class. Unreliable predictions can be filtered out by removing images with the low-confidence pseudo-labels (i.e. when confidence $conf(\cdot) < thresh$). We define the confidence of the predicted segmentation mask as the negative mean entropy of the pixel labels in the mask. 

Every next round, we first (a) retrain the model using jointly ground truth labels and confident pseudo-labels; and then (b) update the pseudo-labels for all unlabeled data using the new model. It is crucial to reset model weights before every round of self-training not to accumulate errors in pseudo-labels during multiple rounds \cite{self_training_error_amp}. 

To further improve the robustness of the generated pseudo-labels and prevent over-fitting to the errors of the sole model, we jointly train an ensemble of CNNs with different backbone architectures. In this case, the pseudo-labels are produced by averaging the predictions of all models in the ensemble. And every next round each model in the ensemble utilizes the confident knowledge of the entire ensemble from the previous round expressed and aggregated in the pseudo-labels. We summarize the full self-training procedure in Algorithm~\ref{alg:self_training} and visualize it in Fig.~\ref{fig:self_training_pipeline}.

\begin{algorithm}[t]
    \SetAlgoLined
    \SetKwInOut{Input}{Input}
    
    \Input{Labeled data $\mathcal{D}_{gt} = (X, Y)$, where $X$ is the set of images and $Y$ is their corresponding ground truth labels; unlabeled images $\Tilde{X}$; number of training epochs $T$; number of self-training rounds $K$; model $\phi(\cdot, \theta)$ with learnable parameters $\theta$.}
    \vspace{8pt}
    
    $\mathcal{D} \gets \mathcal{D}_{gt}$; \Comment{initialize the training set}\\
    \For{$k \gets 1$ to $K$}{               
        Initialize $\theta$ using a pretrained Imagenet model\;
        Train model $\phi(\cdot, \theta)$ on $\mathcal{D}$ for $T$ epochs\;
        Predict pseudo-labels $\hat{Y}$ for each element in $\hat{X}$\;
        $\mathcal{D}_{pseudo} \gets (\hat{X}, \hat{Y})$\;
        Remove images with low-confidence pseudo-labels from $\mathcal{D}_{pseudo}$\;
        
        $\mathcal{D} \gets \mathcal{D}_{gt} \cup \mathcal{D}_{pseudo}$;
    }
    
    \KwResult{Model parameters $\theta$.}
\caption{The proposed self-training procedure}
\label{alg:self_training}
\end{algorithm}

\subsection{Network architecture}\label{sec:architecture}
We start building our networks inspired by seminal U-Net architecture \cite{unet}, which has an encoder, a decoder and skip connections between encoder and decoder blocks with similar spatial resolution. However, training the encoder from scratch is difficult given a limited amount of labeled data. Hence, we opt to use an Imagenet pretrained CNN as the backbone for the encoder \cite{unet_pretrained}.
In particular, we build an ensemble of two models: the first one uses ResNet34 \cite{resnet} as the encoder backbone (we will refer to it as U-ResNet34) and the second one with ResNeXt50 \cite{resnext} as the encoder backbone (we will refer to it as U-ResNeXt50).

We propose a number of modifications to the architecture to make it more effective for salt body delineation task. 
We use several types of attention mechanisms in the network. The encoder and decoder consist of repeating blocks separated by down-sampling and up-sampling respectively.
First, we insert concurrent spatial and channel Squeeze \& Excitation modules (scSE) \cite{scse} after each encoder and decoder block. scSE modules can be interpreted as some sort of attention mechanism: they rescale individual dimensions of the feature maps by increasing the importance of informative features and suppressing the less relevant ones. 

Additionally, in the bottleneck block between the encoder and the decoder we use Feature Pyramid Attention module \cite{fpa}, which increases the receptive field by fusing features from different pyramid scales.

Another powerful design decision for exploiting feature maps from different scales is Hypercolumns \cite{hypercolumns}. Instead of using only the last layer of the decoder for prediction of the segmentation mask, we stack the upsampled feature maps from all decoder blocks and use them as the input to the final layer. It allows getting more precise localization and captures the semantics at the same time. To produce the final segmentation mask, we feed Hypercolumns through a $3 \times 3$ convolution followed by the final $1 \times 1$ convolution. 
We present our final network architecture in Fig. \ref{fig:architecture}.

\section{Experiments}

\begin{table*}

\caption{Results and ablation studies. The first section of the table shows the performance of a single U-ResNet34 without any usage of pseudo-labels. The second block shows the quality increase after several self-training rounds for a single U-ResNet34 model. The third block shows results for multiple self-training rounds using the ensemble of U-ResNet34 and U-ResNeXt50 networks which achieves state-of-the-art performance. Finally, our best model is compared with another approach presented in \cite{kaggle_salt_report_Karchevskiy} on the same dataset.}

\centering

\begin{tabular}{lccc}

\toprule
Method &
Private test mAP &
Public test mAP &
Private LB place \\
\toprule

\tabsection{Our U-ResNet34}{Round 1 ablation studies}
 Single best snapshot & $0.8682$ & $0.8431$ & $200$\\ 
 + TTA & $0.8739$ ($+0.6\%$) & $0.8498$ & $144$ \\
 + Multiple snapshots & $0.8777$ ($+0.4\%$) & $0.8552$ & $99$ \\
 + Multiple folds & $0.8834$ ($+0.6\%$) & $0.8629$ & $61$ \\ 
 + Train 200 epochs more & $0.8845$ ($+0.1\%$) & $0.8644$ & $51$ \\ 
 
\tabsection{Our U-ResNet34}{}

Round 1 & $0.8834$ & $0.8629$ & $61$ \\
Round 2 & $0.8898$ ($+0.6\%$) & $0.8719$ & $20$ \\
Round 3 & $0.8915$ ($+0.2\%$) & $0.8715$ & $12$ \\
Ensemble of Rounds 2 and 3 & $0.8917$ ($+0.1\%$) & $0.8727$ & $10$ \\

\tabsection{Our U-ResNet34 + U-ResNeXt50}{}

Round 1 & $0.8853$ & $0.8677$ & $46$ \\
Round 2 & $0.8919$ ($+0.7\%$) & $0.8748$ & $10$ \\
Round 3 & $0.8953$ ($+0.4\%$) & $0.8759$ & $5$ \\

Ensemble of Rounds 2 and 3 & $\boldsymbol{0.8964}$ ($+0.1\%$) & $\boldsymbol{0.8766}$ & $\boldsymbol{1}$ \\

\midrule
\midrule
\cite{kaggle_salt_report_Karchevskiy} & $0.8880$ & $0.8663$ & $27$ \\

\bottomrule

\end{tabular}

\label{tab:results}

\end{table*}

\subsection{Dataset: TGS Salt Identification Challenge}

TGS Salt Identification Challenge is a Machine Learning competition on a Kaggle platform \cite{kaggle_salt_competition}. The data for this competition represents 2D image slices of 3D view of earth's interior. It was collected using reflection seismology method (similar to X-ray, sonar, and echolocation).
For this reason, input data is a set of single-channel grayscale images showing the boundaries between different rock types at various locations chosen at random in the subsurface. For the competition purposes, large-size images were transformed into $101 \times 101$ pixel crops by the organizers. Further, each pixel is classified as either salt or sediment and binary masks are provided. To visualize the data we assembled a mosaic using the several small patches from the dataset (see Fig. \ref{fig:data_example_mosaic}). The goal of the competition is to segment regions that contain salt. Note that if the 101x101 image contains all the salt pixels, it is treated as an empty mask in the data. Such peculiarity is explained by the organizers as they are more interested in segmenting salt deposit boundaries instead of full-body salt.

The whole dataset has been split into three parts: train, public test, and private test. The train set consists of $4000$ images together with binary masks and is used for models developing. The public test set has around $6000$ images and is used for evaluating the models during the competition. Lastly, private test set has around $12000$ images and is used to determine the final competition standings. Overall, the test dataset contains $18000$ unlabeled images (public + private test) which we can use for self-training.

To track the local quality of the models and prevent overfitting we used 5-fold cross-validation. Thus, every model is trained five times (one per fold). 

\subsubsection{Evaluation metric}

The metric used in this competition is defined as the mean average precision at $10$ different intersection over union (IoU) thresholds $t=(0.50, 0.55, \dots, 0.95)$.
The IoU of a predicted set of salt pixels and a set of true salt pixels is calculated as: 
\begin{equation}
IoU(A,B) = \frac{A \cap B}{ A \cup B}.
\end{equation}

Let $Y$ be a ground truth set of pixels and $Y'$ be a set of pixels predicted by a model. At each threshold $t$, a precision value is calculated based on the following rules:

\begin{equation}
    P(t) = \begin{cases} 0, & \mbox{if } |Y| = 0 \mbox{ and } |Y'| > 0 \\
    0, & \mbox{if } |Y| > 0 \mbox{ and } |Y'| = 0 \\
    1, & \mbox{if } |Y| = 0 \mbox{ and } |Y'| = 0 \\
    IoU(Y, Y') > t, & \mbox{if } |Y| > 0 \mbox{ and } |Y'| > 0 \end{cases}
\end{equation}

Then, the average precision of a single image is calculated as the mean of the above precision values at each IoU threshold: 
\begin{equation}
    AP = \frac{1}{10} \sum_{i=1}^{10}P(t_i).
\end{equation}
The final evaluation score (mAP) is calculated as the mean taken over the individual average precisions of each image in the test dataset.

\subsection{Implementation details}

We employ an ensemble of two U-Nets with Imagenet-pretrained encoder backbones: U-ResNet34 and U-ResNeXt50. The output of the ensemble is the average of the predictions of two models in the ensemble.

All images are resized to the size of $202 \times 202$ pixels and then padded to the size of $256 \times 256$ pixels. 
We do $K=3$ rounds of self-training and $T=200$ training epochs per round. Increasing the number of rounds did not lead to significant improvements of the results.
We use cosine annealing learning rate policy \cite{cosine_annealing} resetting the learning rate every $50$ epochs (cf. loss spikes in Fig.~\ref{fig:train_progress_plot} every $50$ epochs). The learning rate starts from $0.001$ and decays to $0.0001$ every cycle. 

Model weights are "warmed-up" using binary cross-entropy loss during the first $50$ epochs.
After that, we minimize Lovasz loss function \cite{lovasz} for $150$ epochs, which allows a direct optimization of the IoU metric. The warm-up phase is necessary because we noticed that the network gets stuck in a very bad local optimum when the Lovasz loss is used from the very beginning.

Additionally, to get a more robust ensemble at the end of every round we average the predictions of 4 snapshots, which are saved every 50 epochs. When predicting pseudo-labels for all unlabeled data we do not remove the low-confidence predictions (i.e. $thresh=-\inf$) and use all $18000$ pseudo-labeled images for training in the next round. We noticed that this strategy yielded better results than using only confident pseudo-labels.

During the first self-training round, we train the ensemble on the provided $4000$ labeled images and generate $18000$ pseudo-labels for unlabeled images.
At rounds 2 and 3 we train the network for $T$ epochs solely on the pseudo-labeled data and then fine-tune for another $T$ epochs on the ground truth labeled training images. During initial experiments, we observed that jointly training on the ground-truth labeled images and pseudo-labeled images led to inferior results.

After each stage, we obtain 4 network snapshots for each of 5 folds giving 20 snapshots in total for a single network architecture. Since we use an ensemble of U-ResNet34 and U-ResNeXt50, it results in 40 models in total which are combined together for inference using the average voting.

For the final prediction on the test set, we use an ensemble of Round 2 and Round 3 models, which gives the best performing results on the public and private test sets (see Tab.~\ref{tab:results}).

To ensure the reproducibility, we will release the source code for our approach after the acceptance of the paper.

\subsection{Results}
We now compare our approach to the other state-of-the-art approaches.
The detailed results are presented in Tab.~\ref{tab:results}. We evaluate using 3 metrics: private test mAP; public test mAP; place the model achieves on the private leaderboard (LB). 

The table is split into three sections. The first section shows the results of the single U-ResNet34 model without the usage of pseudo-labels (i.e. Round 1 only). The second section ("Our U-ResNet34") shows results for 3 rounds of self-training using our U-ResNet34 model only (no ensemble used).
And the third section ("Our U-ResNet34 + U-ResNeXt50") shows the results for 3 rounds of self-training using the ensemble of U-ResNet34 and U-ResNeXt50.

Training U-ResNet34 for $200$ epochs gives $0.8834$ mAP on private test. If we continue training the same model for another $200$ epochs, it gives only a minor improvement by $0.1\%$.

However, the proposed self-training procedure allows to further improve the score using the unlabeled data while regular training does not help anymore. Round 2 of self-training significantly improves the performance: private test mAP score is increased by $0.6\%$ bringing the model $41$ positions up the leaderboard. Round 3 further improves the mAP score on the private test by $0.2\%$ and moves us to the $12$-th position on the leaderboard. This time the improvement is not so large as after Round 2, nevertheless it shows that applying multiple self-training rounds allows the model to iteratively increase the quality. Finally, a simple average of Round 2 and Round 3 models gives an extra $0.1\%$ performance boost and brings us to the $10$ place on the leaderboard (see "Ensemble of Round 2 and Round 3" in the second section of Tab.~\ref{tab:results}). Fig.~\ref{fig:train_progress_plot} shows the validation loss and mAP during different rounds of self-training. We observe that the model achieves better validation score every consequent round of self-training.

Our ensemble of U-ResNet34 and U-ResNeXt50 achieves the top-1 score on the private and public leaderboards showing the state-of-the-art performance on this dataset after two rounds of self-training. It has mAP score $0.8964$ on the private LB (see "Ensemble of Round 2 and Round 3" in the third section of Tab.~\ref{tab:results}).
For comparison, this ensemble surpasses the approach from the $27$-th position described in \cite{kaggle_salt_report_Karchevskiy} by $0.9\%$.

\subsection{Ablation study}
In this section, we investigate improvements that can be gained using only one model architecture. The results of the ablations studies are reported in the first section of the Tab.~\ref{tab:results}.

We start with a single best snapshot of U-ResNet34 model which yields $0.8682$ private test mAP ($200$-th place on the private leaderboard).
The first idea is to use Test Time Augmentations (TTA): instead of predicting on a single test image, we average predictions on the original test image and its horizontal flip. Such an approach gives $0.6\%$ performance boost almost for free.

The next idea is to utilize multiple snapshots. As was shown in the previous section, the cosine annealing learning rate schedule allows us to obtain multiple local optima in a single training loop. We can create an ensemble of all the snapshots instead of using only the latest snapshot. Such a method gives another $0.4\%$ performance improvement.

Finally, we can further increase the diversity of the models training them on different data subsets. The most obvious choice, in this case, is to use $k$-fold data split and train $k$ different models. This simple idea gives another substantial improvement of $0.6\%$ mAP score relative to the previous one. It corresponds to the 40 positions increase on the private leaderboard.

\section{Conclusion}
We introduced an iterative self-training approach for semantic segmentation which can be effectively used in the limited labeled data setup by using unlabeled data to boost the model performance. Moreover, we designed a sophisticated network architecture for the task of salt body delineation and evaluated the proposed approach on a real-world salt body delineation dataset -- TGS Salt Identification Challenge \cite{kaggle_salt_competition}. Our approach shows the best performance in the TGS Salt Identification Challenge \cite{kaggle_salt_competition} reaching the top-1 position on the leaderboard among the $3234$ competing teams, which proves its effectiveness for the task.

\section*{Acknowledgements}
We would like to thank Pavel Yakubovskiy for the segmentation models zoo in Keras \cite{qubvel} and authors of the Albumentations library \cite{buslaev2018albumentations} for fast and flexible image augmentations. Also, special thanks to Open Data Science community \cite{ods} for many valuable discussions and educational help in the growing field of machine learning.

\bibliographystyle{splncs04}
\bibliography{egbib}

\end{document}